\documentclass[runningheads]{llncs}
\usepackage{graphicx}
\usepackage{multirow}
\usepackage{anyfontsize}
\begin{document}
\title{AttoNets: Compact and Efficient Deep Neural Networks for the Edge via Human-Machine Collaborative Design
\thanks{The authors would like to thank Akif Kamal, Francis Li, Stanislav Bochkarev, David Dolson, Sidharth Somanathan, and Yihan Duan at DarwinAI Corp. for their support.}}
\titlerunning{AttoNets: Deep Neural Networks for the Edge}
%
\author{
    Alexander Wong\inst{1, 2, 3}  
    \and
    Zhong Qiu Lin\inst{1, 2, 3} 
    \and
    Brendan Chwyl\inst{3}   
}
\authorrunning{A. Wong et al.}
%
\institute{
    University of Waterloo, Waterloo, ON, Canada
    \and
    Waterloo Artificial Intelligence Institute, Waterloo, ON, Canada
    \and
    DarwinAI Corp., Waterloo, ON, Canada
}
\maketitle              
\begin{abstract}
While deep neural networks have achieved state-of-the-art performance across a large number of complex tasks, it remains a big challenge to deploy such networks for practical, on-device edge scenarios such as on mobile devices, consumer devices, drones, and vehicles. There has been significant recent effort in designing small, low-footprint deep neural networks catered for low-power edge devices, with much of the focus on two extremes: hand-crafting via design principles or fully automated network architecture search.  In this study, we take a deeper exploration into a human-machine collaborative design approach for creating highly efficient deep neural networks through a synergy between principled network design prototyping and machine-driven design exploration.  The efficacy of human-machine collaborative design is demonstrated through the creation of AttoNets, a family of highly efficient deep neural networks for on-device edge deep learning. Each AttoNet possesses a human-specified network-level macro-architecture comprising of custom modules with unique machine-designed module-level macro-architecture and micro-architecture designs, all driven by human-specified design requirements.  Experimental results for the task of object recognition showed that the AttoNets created via human-machine collaborative design has significantly fewer parameters and computational costs than state-of-the-art networks designed for efficiency while achieving noticeably higher accuracy (with the smallest AttoNet achieving $\sim$1.8\% higher accuracy while requiring $\sim$10$\times$ fewer multiply-add operations and parameters than MobileNet-V1).  Furthermore, the efficacy of the AttoNets is demonstrated for the task of instance segmentation and object detection, where an AttoNet-based Mask R-CNN network was constructed with significantly fewer parameters and computational costs ($\sim$5$\times$ fewer multiply-add operations and $\sim$2$\times$ fewer parameters) than a ResNet-50 based Mask R-CNN network.

\end{abstract}

\section{Introduction}
\label{introduction}
Deep learning~\cite{lecun2015deep} has garnered considerable attention, with deep neural networks achieving incredible state-of-the-art performance across a large variety of complex tasks, particularly in the domain of visual perception, ranging from image categorization~\cite{krizhevsky2012imagenet}, object detection~\cite{fasterrcnn,liu2016ssd},  and image segmentation~\cite{lin2017refinenet,chen2018deeplab,he2017mask}.  Despite these tremendous successes over recent years, the increasing complexities of deep neural networks pose a tremendous challenge in the widespread adoption and deployment for practical, on-device edge scenarios such as on mobile devices, consumer devices, drones, and vehicles where computational, memory, bandwidth, and energy resources are scarce.  As such, there has been significant recent effort in designing small, low-footprint deep neural networks catered for low-power edge devices.

To tackle the challenge of adoption and deployment of deep learning on the edge, there has been an increased focus on exploring strategies for producing highly efficient deep neural networks in a number of different directions. One direction that has been explored is to first design a complex but powerful deep neural network and then perform precision reduction~\cite{Jacob,Meng2017,courbariaux2015binaryconnect}, where the internal components of the designed deep neural network are represented at fixed-point or integer precision precision~\cite{Jacob}, 2-bit precision~\cite{Meng2017}, or even 1-bit precision~\cite{courbariaux2015binaryconnect}.  One limitation of such an approach is that there can be large trade-offs between precision and accuracy, as well as require special reduced-precision acceleration support on the target processor.   In a similar vein, researchers have also explored the strategy of first designing a complex but powerful deep neural network and then leveraging model compression~\cite{han2015deep,hinton2015distilling,projectionnet}, where traditional data compression methods such as weight thresholding, hashing, and Huffman coding, as well as teacher-student strategies involving a larger teacher network training a smaller student network.

Another key direction in the design of highly efficient deep neural networks for edge and mobile scenarios that have been heavily explored in recent years focuses on the fundamental design of deep neural networks itself by introducing architectural design principles to achieve more efficient deep neural network architectures~\cite{MobileNetv1,MobileNetv2,SqueezeNet,SquishedNets,TinySSD,ShuffleNetv1,ShuffleNetv2,ResNet}.  For example, He et al.~\cite{ResNet} introduced a bottleneck module-level macro-architecture, where a 1$\times$1 convolutional layer is first used to reduce the dimensionality entering a 3$\times$3 convolutional layer, with a second 1$\times$1 convolutional layer following the 3$\times$3 convolutional layer to restore dimensionality.  This thus creates a bottleneck at the 3$\times$3 convolutional layer with lower input/output dimensionality and enables deeper architectures to be constructed at a considerably lower computational and memory complexity.
Iandola et al.~\cite{SqueezeNet} introduced an alternative bottleneck module-level macro-architecture in the form of a squeeze-expand macro-architecture, which consists of a 1$\times$1 'squeeze' convolutional layer followed by an 'expand' convolutional layer comprised of a combination of 1$\times$1 and 3$\times$3 convolutional filters.  Howard et al.~\cite{MobileNetv1} leveraged the notion of depthwise separable convolutions to reduce both computational and memory requirements, where the conventional convolutional layer is replaced by a factorized variant composed of (i) a depthwise convolutional layer where a single convolutional filter is applied to each input channel, followed by (ii) a 1$\times$1 convolutional layer to perform pointwise convolutions for computing new features via linear combinations of input channels. Furthermore, Sandler et al.~\cite{MobileNetv2} leveraged not just depthwise separable convolutions but also introduced thin linear bottleneck layers for further improving the balance between modeling performance and efficiency.

More recently, an interesting new direction taken by researchers is the exploration of fully automated network architecture search strategies for algorithmically designing highly efficient deep neural networks for edge and mobile scenarios.  Notable strategies include MONAS~\cite{MONAS} and ParetoNASH~\cite{ParetoNASH}, where the search of efficient deep neural networks was formulated as a multi-objective optimization problem, with the objectives including model size and accuracy and solved via reinforcement learning and evolutionary algorithm, respectively.  Finally, in Tan et al.~\cite{MNAS}, a network architecture search strategy based on reinforcement learning was proposed to search for highly efficient deep neural network architectures that provides the best trade-off between accuracy and latency for specific mobile hardware.

As can be seen, much of the recent focus in the direct design of highly efficient deep neural networks for edge and mobile scenarios falls into one of two extremes: (i) hand-crafting via design principles, or (ii) fully automated network architecture search.  In this study, we take an interesting new direction and perform a deeper exploration into the efficacy of a human-machine collaborative design approach for creating highly efficient deep neural networks through a synergy between human-driven principled network design prototyping and machine-driven design exploration. This more balanced approach takes full advantage of human experience and creativity with the meticulousness and raw speed of a machine.  The efficacy of human-machine collaborative design is demonstrated through the creation of AttoNets, a family of highly efficient deep neural networks for on-device edge deep learning.

This paper is organized as follows.  Section 2 describes in detail the human-machine collaborative design strategy for creating highly efficient deep neural networks for edge and mobile scenarios. Section 3 presents the resulting module-level macro-architecture and micro-architecture designs of the family of AttoNets created via the proposed design strategy.  Experimental results for the tasks of object recognition and instance segmentation and object detection are presented and discussed in Section 4, while conclusions are drawn in Section 5.
\section{Methods}
\label{method}
In this study, we explore a human-machine collaborative design strategy for creating highly-efficient deep neural networks, which comprises of two key design phases: i) principled network design prototyping, and ii) machine-driven design exploration.  First, design principles are leveraged to design the network-level macro-architecture and construct a human-specified initial design prototype catered towards visual perception. Second, machine-driven design exploration is performed based not only on the human-specified initial design prototype, but also a set of human-specified design requirements to explore module-level macro-architecture and micro-architecture designs and generate a set of alternative highly-efficient deep neural networks appropriate to the problem space.
By taking a more balanced approach, one can better leverage the strength of human designers in their ability to come up with creative, high-level solutions as well as design constraints and requirements through human ingenuity and experience, all the while leveraging the raw speed and meticulousness of machines in exploring numerous low-level, design factors in a detail-oriented manner within the human-specified design constraints and requirements.  Details of these two key design phases are described in detail below.
\subsection{Principled network design prototyping}
\begin{figure*}[t]
    \centering
    \includegraphics[width=\textwidth]{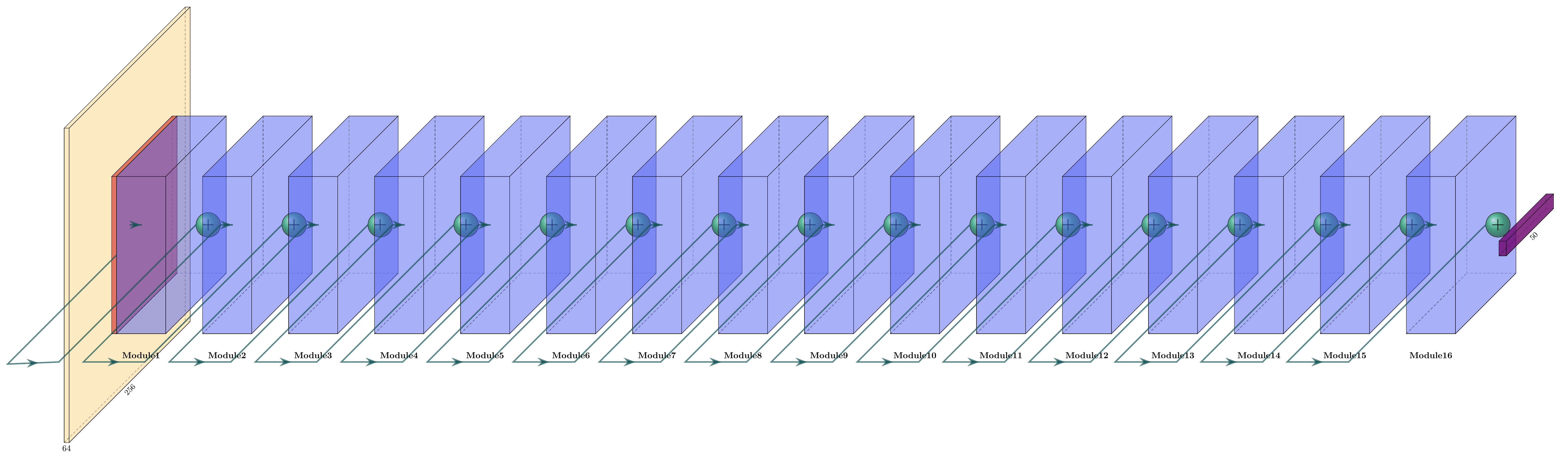}
    \vspace{-0.3in}
    \caption{The human-specified initial design prototype at the principled network design prototyping phase.}
    \label{fig:prototype}
    \vspace{-0.2in}
\end{figure*}

The first phase of the presented human-machine collaborative design  strategy for creating highly-efficient deep neural networks for edge and mobile scenarios is to construct a human-specified initial design prototype by leveraging a number of human-driven design principles.  Rather than focusing our efforts on leveraging design principles centered around efficiency, which in research literature have largely focused on module-level macro-architecture design considerations, we instead focus our human design efforts on leveraging broader network-level macro-architecture design principles centered around modeling performance when designing the initial design prototype.  While this may sound counter-intuitive given our goal is to create highly-efficient deep neural networks, the intuition behind this human focus primarily on modeling performance as opposed to primarily on efficiency is that machines are considerably more capable at low-level design exploration than humans, and as such low-level design exploration for efficiency is best left the primary focus of the machine-driven design exploration phase while human effort is focused more on designing the high-level network infrastructure for achieving high modeling accuracy.

In this particular study, we leverage two human-driven design principles that have been shown in research literature for producing high-performance deep convolutional neural networks for the visual perception task of object recognition, where the goal is to recognize the category in which the contents of a given scene belongs to from a set of object categories (e.g., in the popular ImageNet~\cite{ImageNet} dataset, categories include cellphone, water bottle, monitor, etc.).  To achieve high performance for the task of object recognition, the first well-known design principle in research literature we will leverage is to greatly increase the depth of the underlying deep network architecture, as demonstrated by a variety of different network architectures proposed in literature~\cite{vgg,inceptionv3,tiramisu,ResNet}.  Extending the network architecture depth enables the learning of deeper feature embeddings that can better characterize the complex compositional nature of natural images, formed by visual content at different levels of abstraction and scale.

One of the biggest challenges with leverage very deep network architectures is the difficulty in training such networks, with the difficulty increasing as the depth of the network increases. To mitigate this problem and enable deeper architectures to be built and trained to higher accuracies, the second design principle we will leverage is to incorporate \textit{shortcut connections} within the network architecture.  Introduced by He et al.~\cite{ResNet}, such shortcut connections embedded within the  networks results in a residual architecture that makes it easier for iterative optimization methods such as stochastic gradient descent to drive towards better solutions, even as depth increases in the deep neural network architecture.

Driven by these two human-driven design principles that have been shown to be effective at creating high-performance deep convolutional neural networks, the human-specified initial design prototype possesses a  deep convolutional neural network architecture with a modular design, where the intermediate representation layers are encapsulated within network modules, with shortcut connections between the network modules to enable easier training.  Here, taking inspiration from~\cite{ResNet}, the number of modules in the initial design prototype was set to 16 network modules, which gives enough flexibility for machine-driven design exploration while allowing for sufficient modular depth for achieving strong modeling performance.  After the network modules of intermediate representation layers, we specify an average pooling layer in the initial design prototype, followed by a dense layer. Finally, we specify a softmax layer as the output of the initial design prototype to indicate which of the categories the input image belongs to.  The human-specified initial design prototype is shown in Fig.~\ref{fig:prototype}.  As mentioned earlier, the actual macro-architecture and micro-architecture designs of the individual network modules are left open-ended for the machine-driven design exploration phase to design in an automatic fashion based on the given dataset as well as the set of human-specified design requirements and constraints that cater to deployment for edge scenarios where low computational and memory complexity are desired.
\subsection{Machine-driven design exploration}
The second key phase in the presented human-machine collaborative design strategy for creating highly-efficient deep neural networks for edge and mobile scenarios is to perform machine-driven design exploration to explore module-level macro-architecture and micro-architecture designs based on the human-specified initial design prototype as well as the task and data at hand.  This exploration results in the generation of a set of alternative highly-efficiency deep neural networks, guided by human insights and experience.  In addition, to further exploit the notion of human-machine collaborative design, the machine-driven design exploration process must also take into consideration a set of human-specified design requirements and constraints.  This ensures that the generated deep neural networks produced by machine-driven design exploration are well-suited for on-device object recognition for edge and mobile scenarios.

In this study, we achieve machine-driven design exploration in the form of generative synthesis~\cite{Wong2018} for very fine-grain macro-architecture and micro-architecture design exploration while learning from human-specified design requirements and constraints as well as human-specified network-level macro-architecture designs.  The highly flexible nature of generative synthesis makes it very well-suited for human-machine collaborative design of highly-efficient deep neural networks.

The concept of generative synthesis can be briefly described as follows. The overarching objective of generative synthesis is to learn a generator $\mathcal{G}$ that, given a set of seeds $S$, can generate deep neural networks $\left\{N_s|s \in S\right\}$ that maximize a universal performance function $\mathcal{U}$ while satisfying quantitative requirements defined by an indicator function $1_r(\cdot)$.  This can be formulated as the following constrained optimization problem,

\begin{equation}
\mathcal{G}  = \max_{\mathcal{G}}~\mathcal{U}(\mathcal{G}(s))~~\textrm{subject~to}~~1_r(\mathcal{G}(s))=1,~~\forall s \in S.
\label{optimization}
\end{equation}

In this study, when leveraged for machine-driven design exploration, one can arrive to an approximate solution $\hat \mathcal{G}$ to the constrained optimization problem posed in Eq.~\ref{optimization} in a progressive fashion, where we initialize the generator (i.e., $\hat \mathcal{G}_0$) based on the human-specified initial design prototype $\varphi$, $\mathcal{U}$, and $1_r(\cdot)$, and construct a number of successive generators (i.e., $\hat \mathcal{G}_1$, $\hat \mathcal{G}_2$, $\hat \mathcal{G}_3$, $\ldots$, etc.) via iterative optimization, with each constructed generator $\hat \mathcal{G}_k$ achieving a higher $\mathcal{U}$ than its predecessor generators (i.e., $\hat \mathcal{G}_1$, $\ldots$, $\hat \mathcal{G}_{k-1}$, etc.) while still satisfying $1_r(\cdot)$.  We take full advantage of this interesting phenomenon by leveraging this set of generators to synthesize a family of highly-efficient deep neural networks that satisfies these requirements but with different trade-offs between modeling accuracy and efficiency.

In this study, we leverage the performance function $\mathcal{U}$ introduced in~\cite{Wong2018_Netscore}, which quantifies the balance between modeling performance, computational complexity (which is quantified here by the number of multiply-add operations required for inference), and architectural complexity (which is quantified here by the number of parameters in the deep neural network), thus making it well-suited for driving generative synthesis towards highly-efficient deep neural networks that are well-suited for the computational and memory constraints of edge and mobile scenarios.  Here, we set $\alpha=2, \beta=\gamma=0.5$ as originally proposed in~\cite{Wong2018_Netscore}.  Furthermore, we configure the indicator function $1_r(\cdot)$ such that the top-1 validation accuracy $\geq$ 65\% on ImageNet$_{50}$, a dataset introduced by Fang et al.~\cite{ImageNet50} for evaluating performance of deep neural networks for on-device mobile vision applications.  This indicator function was chosen for this study such that all generated deep neural networks during the machine-driven design exploration process via generative synthesis have a modeling performance that is at least on par with existing state-of-the-art highly-efficient deep convolutional neural networks designed specifically for mobile and edge scenarios such as MobileNet-V1~\cite{MobileNetv1} and ShuffleNet-V2~\cite{ShuffleNetv2}.

\section{AttoNet Architectural Designs}
\label{design}
In this section, we will first describe the module-level macro-architecture designs of the machine-designed custom modules resulting from the machine-driven design exploration phase of the human-machine collaborative design strategy.  Next, we will describe in detail the unique micro-architecture designs created during the machine-driven design exploration phase for each machine-designed custom module within each of the four constructed deep neural networks in the family of AttoNets.
\subsection{Module-level macro-architectures}
\begin{figure*}[t]
    \centering
    \label{fig:attonet}
    \includegraphics[width=1\textwidth]{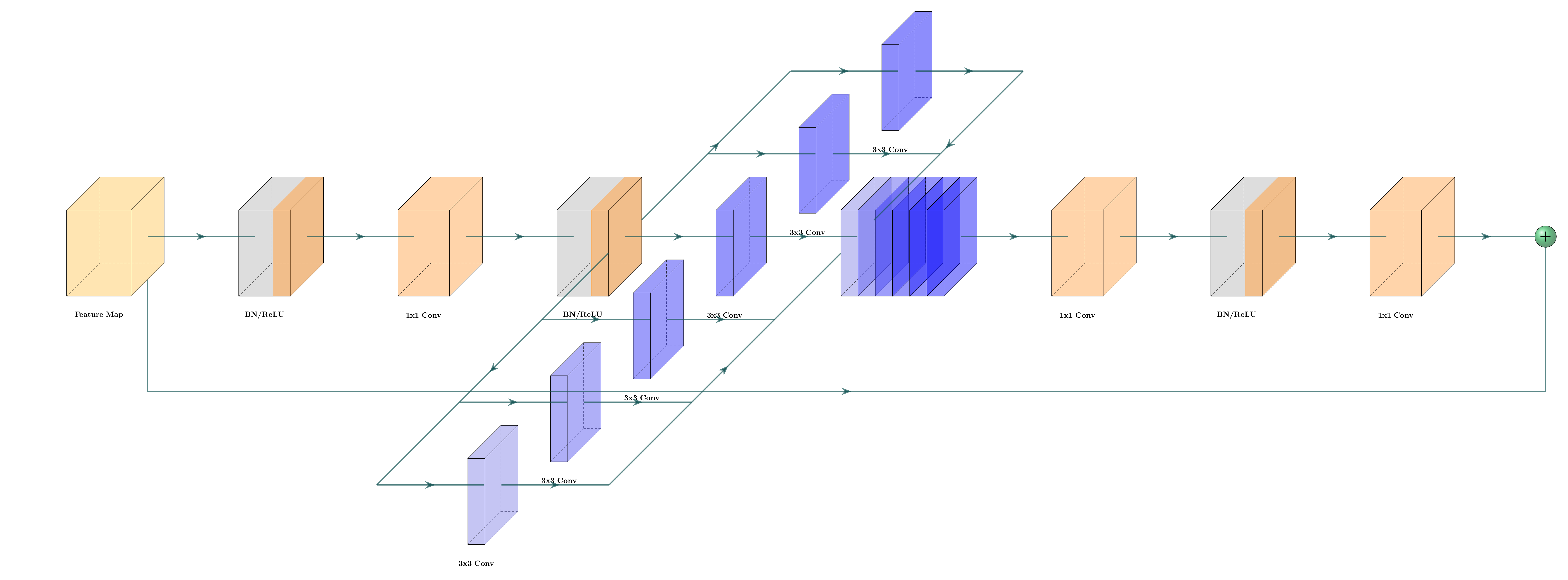}\\
        \vspace{-0.1in}
        ~\\
    AttoNet Module Type A\\
    \vspace{-0.1in}
    ~\\
    \includegraphics[width=1\textwidth]{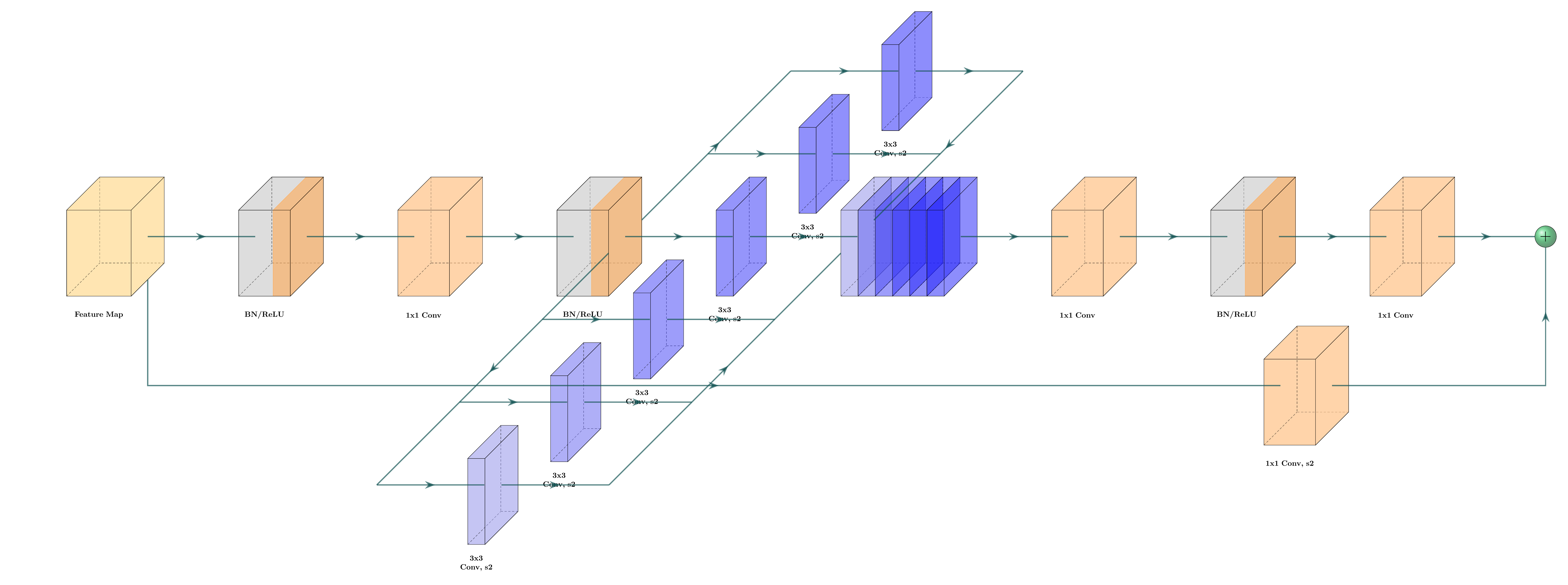}\\
    \vspace{-0.1in}
    (b) AttoNet Module Type B
    \caption{Two module-level macro-architecture designs of the machine-designed custom modules resulting from the machine-driven design exploration phase.  Notable differences in AttoNet Module Type A and AttoNet Module Type B include an additional 1$\times$1 convolutional layer on the shortcut connection in Module Type B, as well as strided convolution groups in Module Type B vs. non-strided convolutional groups in Module Type A. Note that the micro-architecture design (e.g., the number of convolutional groups, size of each group, number of filters of each layer, etc.) is unique for each module within a created network (see Table 1).}
    \vspace{-0.1in}
    \label{fig:attonet}
\end{figure*}
Two module-level macro-architecture designs for the machine-designed custom modules resulting from the machine-driven design exploration phase of the human-machine collaborative design strategy are shown in Fig.~\ref{fig:attonet}.  Notable differences in AttoNet Module Type A and AttoNet Module Type B include an additional 1$\times$1 convolutional layer on the shortcut connection in Module Type B, as well as strided convolution groups in Module Type B vs. non-strided convolutional groups in Module Type A.  Note that the micro-architecture design (e.g., the number of convolutional groups, size of each group, number of filters of each layer, etc.) is unique for each module within a created network (see Table 1).  A number of interesting observations can be made about these unique module-level macro-architecture designs, which we will describe in detail below.
\subsubsection{Pointwise compression.}
\vspace{-0.1in}
A compression layer of 1$\times$1 convolutions is leveraged as the first component of the both Module Type A and Module Type B.  The use of such a compression layer effectively reduces dimensionality into the next layer and thus reduces both the architectural and computational complexity of the custom modules within the network.
\subsubsection{Strided convolutional groups.}
\vspace{-0.1in}
Multiple groups of 3$\times$3 convolutions after the compression layer are leveraged for both Module Type A and Module Type B.  The use of convolutional groups instead of a single convolutional layer enables the output features of the compression layer to be separated into smaller groups and processed independently, thus further reducing the computational complexity and architectural complexity of the modules.  It is important to note that while Module Type A leverages non-strided convolutional groups, Module Type B leverages strided convolution groups instead.  The further use of strided convolutional groups in Module Type B provides another level of information compression by reducing dimensionality into the next layer, which has the effect of further reducing both architectural and computational complexity of the network even more.

\subsubsection{Back-to-back pointwise convolutions.}
\vspace{-0.1in}
Back-to-back 1$\times$1 convolutional layers are leveraged in both Module Type A and Module Type B, with the first 1$\times$1 convolutional layer effectively mixing the outputs of the convolutional groups together to produce features incorporating information from all groups, and the second 1$\times$1 convolutional layer acting as a decompression layer to restore dimensionality in the output of the custom modules.  Therefore, this back-to-back configuration of 1$\times$1 convolutional layers allows for improved performance in the learned feature embedding despite the complexity reductions taken by the compression layer and the convolutional groups.
\subsection{Created AttoNet network architectures}
\begin{table*}[tbh!]
{\fontsize{7}{7}\selectfont
\centering
\caption{Network architectures of proposed family of AttoNets}
\begin{center}
\begin{tabular}{c|c|c|c|c|c}
\hline
~~Module Name~~   & ~~Output size~~            & {AttoNet-A}                                                                                                       & {AttoNet-B}                                                                                                      & {AttoNet-C}                                                                                                     & {AttoNet-D}                                                                                                                       \\ \hline
\hline
Conv          & 112x112                & {7x7, 8, s2}                                                                                                      & {7x7, 8, s2}                                                                                                     & {7x7, 6, s2}                                                                                                    & {7x7, 3, s2}                                                                                                                      \\ \hline
MaxPooling    & \multirow{4}{*}{56x56} & {3x3, 8, s2}                                                                                                      & {3x3, 8, s2}                                                                                                     & {3x3, 6, s2}                                                                                                    & {3x3, 3, s2}                                                                                                                      \\ \cline{1-1} \cline{3-6}
Module1       &                        & \begin{tabular}[c]{@{}c@{}}{[}1x1, 8{]}\\ {[}3x3x4, 32{]}\\ {[}1x1, 16{]}\\ {[}1x1, 176{]} \\ \hline {[}1x1, 176{]}\end{tabular}       & \begin{tabular}[c]{@{}c@{}}{[}1x1, 8{]}\\ {[}3x3x4, 32{]}\\ {[}1x1, 16{]}\\ {[}1x1, 168{]} \\ \hline {[}1x1, 168{]} \end{tabular}      & \begin{tabular}[c]{@{}c@{}}{[}1x1, 7{]}\\ {[}3x3x2, 14{]}\\ {[}1x1, 8{]}\\ {[}1x1, 139{]} \\ \hline {[}1x1, 139{]} \end{tabular}      & \begin{tabular}[c]{@{}c@{}}{[}1x1, 7{]}\\ {[}3x3x4, 28{]}\\ {[}1x1, 8{]}\\ {[}1x1, 112{]} \\ \hline {[}1x1, 112{]} \end{tabular}                          \\ \cline{1-1} \cline{3-6}
Module2       &                        & \begin{tabular}[c]{@{}c@{}}{[}1x1, 16{]}\\ {[}3x3x4,  64{]}\\ {[}1x1, 16{]}\\ {[}1x1, 176{]} \\ \hline Identity \end{tabular}      & \begin{tabular}[c]{@{}c@{}}{[}1x1, 16{]}\\ {[}3x3x4, 64{]}\\ {[}1x1, 8{]}\\ {[}1x1, 168{]}\\ \hline Identity \end{tabular}      & \begin{tabular}[c]{@{}c@{}}{[}1x1, 5{]}\\ {[}3x3x2, 10{]}\\ {[}1x1, 5{]}\\ {[}1x1, 139{]}\\ \hline Identity \end{tabular}            & \begin{tabular}[c]{@{}c@{}}{[}1x1, 5{]}\\ {[}3x3x4, 20{]}\\ {[}1x1, 5{]}\\ {[}1x1, 112{]}\\ \hline Identity \end{tabular}                                \\ \cline{1-1} \cline{3-6}
Module3       &                        & \begin{tabular}[c]{@{}c@{}}{[}1x1, 16{]}\\ {[}3x3x4, 64{]}\\ {[}1x1, 16{]}\\ {[}1x1, 176{]}\\ \hline Identity \end{tabular}           & \begin{tabular}[c]{@{}c@{}}{[}1x1, 16{]}\\ {[}3x3x4, 64{]}\\ {[}1x1, 16{]}\\ {[}1x1, 168{]}\\ \hline Identity \end{tabular}           & \begin{tabular}[c]{@{}c@{}}{[}1x1, 7{]}\\ {[}3x3x2, 14{]}\\ {[}1x1, 6{]}\\ {[}1x1, 139{]}\\ \hline Identity \end{tabular}           & \begin{tabular}[c]{@{}c@{}}{[}1x1, 7{]}\\ {[}3x3x4, 28{]}\\ {[}1x1,65{]}\\ {[}1x1, 112{]}\\ \hline Identity \end{tabular}                               \\ \hline
Module4       & \multirow{4}{*}{28x28} & \begin{tabular}[c]{@{}c@{}}{[}1x1, 40{]}\\ {[}3x3x4, 160, s2{]}\\ {[}1x1, 40{]}\\ {[}1x1, 400{]}\\ \hline {[}1x1, 400, s2{]} \end{tabular}  & \begin{tabular}[c]{@{}c@{}}{[}1x1, 32{]}\\ {[}3x3x4, 128, s2{]}\\ {[}1x1, 24{]}\\ {[}1x1, 368{]} \\ \hline {[}1x1, 368, s2{]} \end{tabular} & \begin{tabular}[c]{@{}c@{}}{[}1x1, 19{]}\\ {[}3x3x2, 38, s2{]}\\ {[}1x1, 14{]}\\ {[}1x1, 328{]} \\ \hline {[}1x1, 328, s2{]}\end{tabular}  & \begin{tabular}[c]{@{}c@{}}{[}1x1, 8{]}\\ {[}3x3x4, 32, s2{]}\\ {[}1x1, 8{]}\\ {[}1x1, 264{]} \\ \hline {[}1x1, 264, s2{]} \end{tabular}                      \\ \cline{1-1} \cline{3-6}
Module5       &                        & \begin{tabular}[c]{@{}c@{}}{[}1x1, 32{]}\\ {[}3x3x4, 128{]}\\ {[}1x1, 24{]}\\ {[}1x1, 400{]}\\ \hline Identity \end{tabular}           & \begin{tabular}[c]{@{}c@{}}{[}1x1, 16{]}\\ {[}3x3x4, 64{]}\\ {[}1x1, 24{]}\\ {[}1x1, 368{]}\\ \hline Identity \end{tabular}           & \begin{tabular}[c]{@{}c@{}}{[}1x1, 13{]}\\ {[}3x3x2, 26{]}\\ {[}1x1, 9{]}\\ {[}1x1, 328{]}\\ \hline Identity \end{tabular}           & \begin{tabular}[c]{@{}c@{}}{[}1x1, 8{]}\\ {[}3x3x4, 32{]}\\ {[}1x1,8{]}\\ {[}1x1, 264{]}\\ \hline Identity \end{tabular}                               \\ \cline{1-1} \cline{3-6}
Module6       &                        & \begin{tabular}[c]{@{}c@{}}{[}1x1, 40{]}\\ {[}3x3x4, 160{]}\\ {[}1x1, 32{]}\\ {[}1x1, 400{]}\\ \hline Identity \end{tabular}           & \begin{tabular}[c]{@{}c@{}}{[}1x1, 24{]}\\ {[}3x3x4, 96{]}\\ {[}1x1, 24{]}\\ {[}1x1, 368{]}\\ \hline Identity \end{tabular}           & \begin{tabular}[c]{@{}c@{}}{[}1x1, 19{]}\\ {[}3x3x2, 38{]}\\ {[}1x1, 8{]}\\ {[}1x1, 328{]}\\ \hline Identity \end{tabular}           & \begin{tabular}[c]{@{}c@{}}{[}1x1, 8{]}\\ {[}3x3x4, 32{]}\\ {[}1x1,8{]}\\ {[}1x1, 264{]}\\ \hline Identity \end{tabular}                               \\ \cline{1-1} \cline{3-6}
Module7       &                        & \begin{tabular}[c]{@{}c@{}}{[}1x1, 40{]}\\ {[}3x3x4, 160{]}\\ {[}1x1, 32{]}\\ {[}1x1, 400{]}\\ \hline Identity \end{tabular}           & \begin{tabular}[c]{@{}c@{}}{[}1x1, 24{]}\\ {[}3x3x4, 96{]}\\ {[}1x1, 32{]}\\ {[}1x1, 368{]}\\ \hline Identity \end{tabular}           & \begin{tabular}[c]{@{}c@{}}{[}1x1, 20{]}\\ {[}3x3x2, 40{]}\\ {[}1x1, 12{]}\\ {[}1x1, 328{]}\\ \hline Identity \end{tabular}           & \begin{tabular}[c]{@{}c@{}}{[}1x1, 8{]}\\ {[}3x3x4, 32{]}\\ {[}1x1, 8{]}\\ {[}1x1, 264{]}\\ \hline Identity \end{tabular}                               \\ \hline
Module8       & \multirow{6}{*}{14x14} & \begin{tabular}[c]{@{}c@{}}{[}1x1, 80{]}\\ {[}3x3x4, 320, s2{]}\\ {[}1x1, 64{]}\\ {[}1x1, 808{]}\\ \hline {[}1x1, 808, s2{]} \end{tabular}  & \begin{tabular}[c]{@{}c@{}}{[}1x1, 64{]}\\ {[}3x3x4, 256, s2{]}\\ {[}1x1, 48{]}\\ {[}1x1, 728{]}\\ \hline {[}1x1, 728, s2{]} \end{tabular}  & \begin{tabular}[c]{@{}c@{}}{[}1x1, 39{]}\\ {[}3x3x2, 78, s2{]}\\ {[}1x1, 24{]}\\ {[}1x1, 628{]}\\ \hline {[}1x1, 628, s2{]} \end{tabular}  & \begin{tabular}[c]{@{}c@{}}{[}1x1, 8{]}\\ {[}3x3x4, 32{]}\\ {[}1x1, 8{]}\\ {[}1x1, 467{]}\\ \hline \multicolumn{1}{l}{{[}1x1, 467, s2{]}} \end{tabular} \\ \cline{1-1} \cline{3-6}
Module9       &                        & \begin{tabular}[c]{@{}c@{}}{[}1x1, 72{]}\\ {[}3x3x4, 288{]}\\ {[}1x1, 72{]}\\ {[}1x1, 808{]} \\ \hline Identity \end{tabular}           & \begin{tabular}[c]{@{}c@{}}{[}1x1, 40{]}\\ {[}3x3x4, 160{]}\\ {[}1x1, 40{]}\\ {[}1x1, 728{]} \\ \hline Identity \end{tabular}           & \begin{tabular}[c]{@{}c@{}}{[}1x1, 29{]}\\ {[}3x3x2, 58{]}\\ {[}1x1, 28{]}\\ {[}1x1, 628{]} \\ \hline Identity \end{tabular}           & \begin{tabular}[c]{@{}c@{}}{[}1x1, 8{]}\\ {[}3x3x4, 32{]}\\ {[}1x1, 8{]}\\ {[}1x1, 467{]} \\ \hline Identity \end{tabular}                               \\ \cline{1-1} \cline{3-6}
Module10      &                        & \begin{tabular}[c]{@{}c@{}}{[}1x1, 80{]}\\ {[}3x3x4, 320{]}\\ {[}1x1, 72{]}\\ {[}1x1, 808{]} \\ \hline Identity \end{tabular}           & \begin{tabular}[c]{@{}c@{}}{[}1x1, 48{]}\\ {[}3x3x4, 192{]}\\ {[}1x1, 48{]}\\ {[}1x1, 728{]}  \\ \hline Identity \end{tabular}           & \begin{tabular}[c]{@{}c@{}}{[}1x1, 37{]}\\ {[}3x3x2, 74{]}\\ {[}1x1, 26{]}\\ {[}1x1, 628{]} \\ \hline Identity \end{tabular}           & \begin{tabular}[c]{@{}c@{}}{[}1x1, 8{]}\\ {[}3x3x4, 32{]}\\ {[}1x1, 8{]}\\ {[}1x1, 467{]} \\ \hline Identity \end{tabular}                               \\ \cline{1-1} \cline{3-6}
Module11      &                        & \begin{tabular}[c]{@{}c@{}}{[}1x1, 72{]}\\ {[}3x3x4, 288{]}\\ {[}1x1, 72{]}\\ {[}1x1, 808{]} \\\hline Identity \end{tabular}           & \begin{tabular}[c]{@{}c@{}}{[}1x1, 56{]}\\ {[}3x3x4, 224{]}\\ {[}1x1, 48{]}\\ {[}1x1, 728{]} \\\hline Identity \end{tabular}           & \begin{tabular}[c]{@{}c@{}}{[}1x1, 37{]}\\ {[}3x3x2, 74{]}\\ {[}1x1, 32{]}\\ {[}1x1, 628{]} \\\hline Identity \end{tabular}           & \begin{tabular}[c]{@{}c@{}}{[}1x1, 8{]}\\ {[}3x3x4, 32{]}\\ {[}1x1, 8{]}\\ {[}1x1, 467{]} \\\hline Identity \end{tabular}                               \\ \cline{1-1} \cline{3-6}
Module12      &                        & \begin{tabular}[c]{@{}c@{}}{[}1x1, 72{]}\\ {[}3x3x4, 288{]}\\ {[}1x1, 64{]}\\ {[}1x1, 808{]}\\\hline Identity \end{tabular}           & \begin{tabular}[c]{@{}c@{}}{[}1x1, 48{]}\\ {[}3x3x4, 192{]}\\ {[}1x1, 40{]}\\ {[}1x1, 728{]}\\\hline Identity \end{tabular}           & \begin{tabular}[c]{@{}c@{}}{[}1x1, 37{]}\\ {[}3x3x2, 74{]}\\ {[}1x1, 27{]}\\ {[}1x1, 628{]}\\\hline Identity \end{tabular}           & \begin{tabular}[c]{@{}c@{}}{[}1x1, 8{]}\\ {[}3x3x4, 32{]}\\ {[}1x1, 8{]}\\ {[}1x1, 467{]}\\\hline Identity \end{tabular}                               \\ \cline{1-1} \cline{3-6}
Module13      &                        & \begin{tabular}[c]{@{}c@{}}{[}1x1, 56{]}\\ {[}3x3x4, 224{]}\\ {[}1x1, 56{]}\\ {[}1x1, 808{]}\\\hline Identity \end{tabular}           & \begin{tabular}[c]{@{}c@{}}{[}1x1, 32{]}\\ {[}3x3x4, 128{]}\\ {[}1x1, 32{]}\\ {[}1x1, 728{]}\\\hline Identity \end{tabular}           & \begin{tabular}[c]{@{}c@{}}{[}1x1, 17{]}\\ {[}3x3x2, 34{]}\\ {[}1x1, 24{]}\\ {[}1x1, 628{]}\\\hline Identity \end{tabular}           & \begin{tabular}[c]{@{}c@{}}{[}1x1, 8{]}\\ {[}3x3x4, 32{]}\\ {[}1x1, 8{]}\\ {[}1x1, 467{]}\\\hline Identity \end{tabular}                               \\ \hline
Module14      & \multirow{3}{*}{7x7}   & \begin{tabular}[c]{@{}c@{}}{[}1x1, 160{]}\\ {[}3x3x4, 640, s2{]}\\ {[}1x1, 128{]}\\ {[}1x1, 952{]}\\\hline {[}1x1, 952, s2{]} \end{tabular} & \begin{tabular}[c]{@{}c@{}}{[}1x1, 112{]}\\ {[}3x3x4, 448, s2{]}\\ {[}1x1, 96{]}\\ {[}1x1, 736{]}\\\hline {[}1x1, 736, s2{]}\end{tabular}  & \begin{tabular}[c]{@{}c@{}}{[}1x1, 67{]}\\ {[}3x3x2, 134, s2{]}\\ {[}1x1, 52{]}\\ {[}1x1, 527{]}\\\hline {[}1x1, 527, s2{]} \end{tabular} & \begin{tabular}[c]{@{}c@{}}{[}1x1, 16{]}\\ {[}3x3x4, 64, s2{]}\\ {[}1x1, 8{]}\\ {[}1x1, 140{]}\\\hline {[}1x1, 140, s2{]} \end{tabular}                      \\ \cline{1-1} \cline{3-6}
Module15      &                        & \begin{tabular}[c]{@{}c@{}}{[}1x1, 120{]}\\ {[}3x3x4, 480{]}\\ {[}1x1, 88{]}\\ {[}1x1, 952{]}\\\hline Identity \end{tabular}           & \begin{tabular}[c]{@{}c@{}}{[}1x1, 72{]}\\ {[}3x3x4, 288{]}\\ {[}1x1, 48{]}\\ {[}1x1, 736{]}\\\hline Identity \end{tabular}           & \begin{tabular}[c]{@{}c@{}}{[}1x1, 46{]}\\ {[}3x3x2, 92{]}\\ {[}1x1, 30{]}\\ {[}1x1, 527{]}\\\hline Identity \end{tabular}           & \begin{tabular}[c]{@{}c@{}}{[}1x1, 8{]}\\ {[}3x3x4, 32{]}\\ {[}1x1, 8{]}\\ {[}1x1, 140{]}\\\hline Identity \end{tabular}                               \\ \cline{1-1} \cline{3-6}
Module16      &                        & \begin{tabular}[c]{@{}c@{}}{[}1x1, 112{]}\\ {[}3x3x4, 448{]}\\ {[}1x1, 88{]}\\ {[}1x1, 952{]}\\\hline Identity \end{tabular}           & \begin{tabular}[c]{@{}c@{}}{[}1x1, 72{]}\\ {[}3x3x4, 288{]}\\ {[}1x1, 56{]}\\ {[}1x1, 736{]}\\\hline Identity \end{tabular}           & \begin{tabular}[c]{@{}c@{}}{[}1x1, 49{]}\\ {[}3x3x2, 98{]}\\ {[}1x1, 31{]}\\ {[}1x1, 527{]}\\\hline Identity \end{tabular}           & \begin{tabular}[c]{@{}c@{}}{[}1x1, 8{]}\\ {[}3x3x4, 32{]}\\ {[}1x1, 8{]}\\ {[}1x1, 140{]}\\\hline Identity \end{tabular}                               \\ \hline
AvgPooling    & 1x1                    & \multicolumn{1}{c|}{7x7, 952}                                                                                                        & \multicolumn{1}{c|}{7x7, 736}                                                                                                       & \multicolumn{1}{c|}{7x7, 527}                                                                                                      & \multicolumn{1}{c}{7x7, 140}                                                                                                                        \\
FC            & 51                     & \multicolumn{1}{c|}{952x51}                                                                                                          & \multicolumn{1}{c|}{736x51}                                                                                                         & \multicolumn{1}{c|}{527x51}                                                                                                        & \multicolumn{1}{c}{140x51}                                                                                                                                                                 \\ \hline
\end{tabular}
\label{tab_AttoNets}
\end{center}
}
\end{table*}

The network-level architecture design, along with the module-level micro-architecture designs of each machine-designed custom module, for each of the four networks in the family of AttoNets produced using the presented human-machine collaborative strategy are summarized in Table~\ref{tab_AttoNets}.  The family of AttoNets possess unique 69-layer deep convolutional neural network architectures, which differs from previously proposed architectures designed for efficiency in edge scenarios.  A number of interesting observations can be made from looking at the individual architectures of the family of produced AttoNets.  First, it is observed that there is considerable inter-network micro-architecture variability, with the module micro-architecture designs of each AttoNet being very different from one another.  Second, it is observed that the overall architectural complexity decreases progressively from AttoNet-A to AttoNet-D, which is consistent with the progressive strategy taken in the machine-driven design exploration phase.  Third, it is observed that there is also considerable intra-network micro-architecture and macro-architecture variability, with each custom module within an AttoNet possessing a unique and very different micro-architecture design compared to the other custom modules, along with exhibiting different module-level macro-architecture designs within the networks as well.  Such diversity in both inter-network and intra-network micro-architecture and macro-architecture variability can only be achieved via fine-grained machine-driven design exploration.

\section{Results and Discussion}
\label{results}
The efficacy of the family of AttoNets created via the presented human-machine collaborative design strategy were quantitatively evaluated for two different popular visual perception tasks that are commonly performed in edge and mobile scenarios: i) object recognition, and ii) instance segmentation and object detection.  The experimental setup and results for the two experiments are presented below.
\subsection{Object Recognition}
The first experiment performed to evaluate the efficacy of the four highly-efficient deep neural networks in the AttoNet family was on the task of object recognition.  More specifically, we evaluate the top-1 accuracy of the presented AttoNets on the ImageNet$_{50}$ dataset~\cite{ImageNet50}, a dataset recently introduced by Fang et al. for evaluating performance of deep neural networks for on-device mobile vision applications derived from the popular ImageNet~\cite{ImageNet} dataset.  For comparison purposes, the results for three state-of-the-art deep neural networks designed for edge and mobile applications (namely, MobileNet-V1~\cite{MobileNetv1}, MobileNet-V2~\cite{MobileNetv2}, and ShuffleNet-V2~\cite{ShuffleNetv2}) are also presented.

\begin{table*}[th!]
\centering
\caption{Performance of efficient deep convolutional neural networks on ImageNet$_{50}$}
\begin{tabular}{c|c|c|c|c}
~~~~~~~Model~~~~~~~                                  & ~~~Top-1~~~             & ~~Mult-Adds (M)~~ & ~~\#Params (M)~~  & ~~NetScore~~       \\ \hline
\hline
\multicolumn{1}{c|}{MobileNet-V1~\cite{MobileNetv1}} & 64.52\%          & 567.5         & 3.26          & 69.71          \\
\multicolumn{1}{c|}{MobileNet-V2~\cite{MobileNetv2}} & 68.68\%          & 299.7         & 2.29          & 75.11          \\
\multicolumn{1}{c|}{ShuffleNet-V2~\cite{ShuffleNetv2}}      & 65.00\%          & 140.1         & 1.32          & 79.85      \\    \hline
\multicolumn{1}{c|}{AttoNet-A}         & \textbf{73.00\%} & 424.8         & 2.97          & 73.53          \\
\multicolumn{1}{c|}{AttoNet-B}         & 71.10\%          & 277.5         & 1.87          & 76.93          \\
\multicolumn{1}{c|}{AttoNet-C}         & 69.60\%          & 139.9         & 1.06          & 81.99          \\
\multicolumn{1}{c|}{AttoNet-D}         & 66.30\%          & \textbf{57.5} & \textbf{0.32} & \textbf{90.21}
\end{tabular}
\vspace{-0.1in}
\label{tab_Results}
\end{table*}

As shown in Table~\ref{tab_Results}, the produced family of AttoNets had noticeably higher accuracies at much smaller sizes and lower computational costs than state-of-the-art deep neural networks designed for edge and mobile applications.  In terms of smallest size, AttoNet-D achieved higher accuracy compared to both MobileNet-V1 and ShuffleNet-V2 ($\sim$\textbf{1.8\%} higher and $\sim$\textbf{1.3\%} higher, respectively ) but requires $\sim$\textbf{10$\times$} fewer parameters and $\sim$\textbf{10$\times$} fewer multiply-add operations than MobileNet-V1 and $\sim$\textbf{4.1$\times$} fewer parameters and $\sim$\textbf{2.4$\times$} fewer multiply-add operations than ShuffleNet-v2.  More interestingly, both AttoNet-B and AttoNet-C achieves higher accuracies compared to the state-of-the-art MobileNet-V2 ($\sim$\textbf{2.4\%} higher and $\sim$\textbf{0.9\%} higher, respectively) but with AttoNet-B requiring $\sim$\textbf{1.2$\times$} fewer parameters and $\sim$\textbf{1.1$\times$} fewer multiply-add operations than MobileNet-V2, and AttoNet-C requiring $\sim$\textbf{2.2$\times$} fewer parameters and $\sim$\textbf{2.1$\times$} fewer multiply-add operations than MobileNet-V2.  In terms of best accuracy, AttoNet-A achieved  $\sim$\textbf{8.5\%} higher accuracy compared to MobileNet-V1 while requiring $\sim$\textbf{1.1$\times$} fewer parameters and $\sim$\textbf{1.3$\times$} multiply-add operations.  In terms of the highest NetScore, AttoNet-D achieved a NetScore that is $\sim$\textbf{20.5} points higher than MobileNet-V1, $\sim$\textbf{15.1} points higher than MobileNet-V2 and $\sim$\textbf{10.4} points higher than ShuffleNet-V2, which demonstrates a strong balance between accuracy, computational complexity, and architectural complexity.
\subsection{Instance segmentation and object detection}
The second experiment performed to evaluate the efficacy of AttoNets created via the presented human-machine collaborative design strategy involved the task of instance segmentation and object detection, where the goal is to segment objects in a given scene and detecting bounding boxes around objects, as well as assigning a category to each segment.  More specifically, we train an AttoNet-based Mask R-CNN network (which we will refer to as Atto-MaskRCNN) on the popular Pascal VOC 2012~\cite{VOC} segmentation dataset and evaluate the mask average precision (which we denote as AP$_{m}$) and the detection (bounding box) average precision (which we denote as AP$_{d}$).  For training, SGD with a momentum of 0.9 was used as described in~\cite{he2017mask}.  In this experiment, ResNet-50 based Mask R-CNN ~\cite{he2017mask} is used for comparison purposes.

\begin{figure*}[tbh]
    \centering
    \includegraphics[width=1\textwidth]{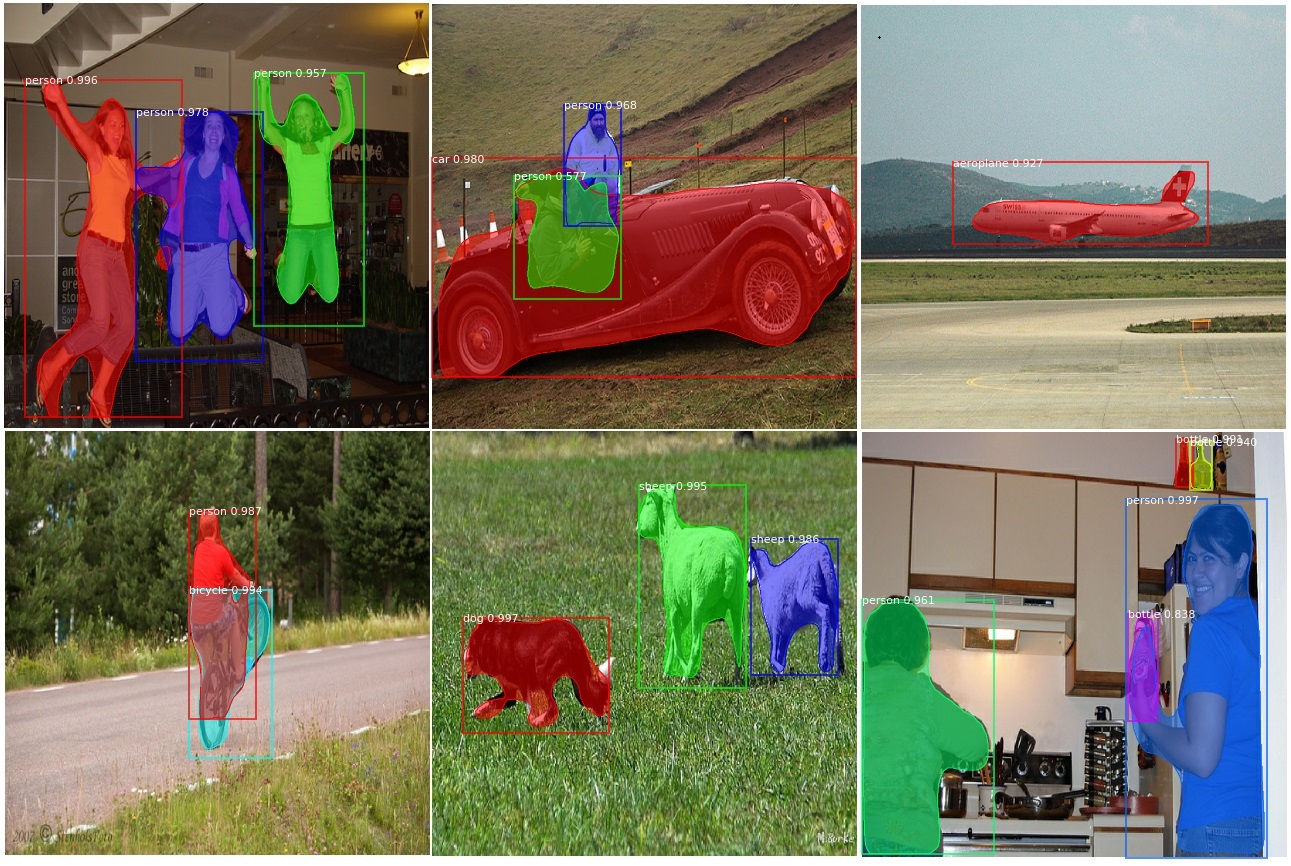}
    \caption{Example images with their corresponding instance segmentations and detected bounding boxes produced by Atto-MaskRCNN from the Pascal VOC 2012 segmentation dataset.  Note that Atto-MaskRCNN achieves good object instance segmentation and detection performance while requiring $\sim$5$\times$ fewer multiply-add operations than ResNet-50 based Mask R-CNN~\cite{he2017mask}.}
    \vspace{-0.2in}
    \label{fig:VOC}
\end{figure*}

It was observed that the produced Atto-MaskRCNN requires $\sim$\textbf{2$\times$} fewer parameters (at 18.3M) and $\sim$\textbf{5$\times$} fewer multiply-add operations (at 195B) than ResNet-50 based Mask R-CNN~\cite{he2017mask}, while still achieving an AP$_{d}$ of \textbf{$\sim$65.5\%} and AP$_{m}$ of \textbf{$\sim$60\%} on the Pascal VOC 2012 segmentation dataset.  Example images with the corresponding instance segmentation and detected bounding boxes produced by Atto-MaskRCNN from the Pascal VOC 2012 segmentation dataset is shown in Fig.~\ref{fig:VOC}.  It can be observed that the produced Atto-MaskRCNN can achieve good object instance segmentation and detection performance despite its significantly reduced computational and architectural complexity.  The results of the two experiments demonstrate that the family of AttoNets were able to achieve state-of-the-art performance while being noticeably smaller and requiring significantly fewer computations, making them well-suited for on-device edge and mobile scenarios.

\vspace{-0.05in}
\section{Conclusions}
\label{conclusions}
\vspace{-0.05in}
In this study, we show that a human-machine collaborative design strategy can be leveraged for creating highly-efficient deep neural networks for on-device edge deep learning scenarios.  By combining human-driven principled network design prototyping with machine-driven design exploration, one can create deep neural networks with unique, machine-designed module-level macro-architecture and micro-architecture designs catered primarily for efficiency within a human-specified network macro-architecture design catered primarily for modeling accuracy.  Experiment results show that family of AttoNets created via the presented human-machine collaborative design strategy can achieve state-of-the-art accuracy while possessing significantly fewer parameters and with significantly lower computational costs.

Future work involves investigate the effectiveness of the presented human-machine collaborative design strategy for a broader range of tasks such as video action recognition, video pose estimation, image captioning, image super-resolution, and image generation.  Furthermore, it would be worth exploring the effect of different permutations of human-specified design requirements on the resulting deep neural networks being created.

%
%
%
\bibliographystyle{splncs04}
\bibliography{bibfile}

\end{document}